\documentclass[runningheads]{llncs}

\usepackage[mobile]{eccv}

\usepackage{eccvabbrv}

\usepackage{graphicx}
\usepackage{booktabs}
\usepackage{wrapfig,lipsum,booktabs}
\usepackage{multirow}
\usepackage{threeparttable}

\newcommand{\OurMethod}{\emph{BrushNet}}
\newcommand{\Benchmark}{\emph{BrushBench}}
\newcommand{\TrainingData}{\emph{BrushData}}

\usepackage{colortbl}
\usepackage[dvipsnames]{xcolor}
\definecolor{Red}{RGB}{192, 0, 0}
\definecolor{Blue}{RGB}{12, 114, 186}
\definecolor{Yellow}{RGB}{218, 169, 20}

\definecolor{HighlightBlue}{RGB}{0, 100, 148}
\definecolor{HighlightRed}{RGB}{230, 57, 70}

\definecolor{LightRed}{HTML}{ffe0e0}
\definecolor{LightBlue}{HTML}{def5ff}
\definecolor{LightYellow}{HTML}{FFF6DB}
\definecolor{LightGreen}{HTML}{eff9f0}

\usepackage[accsupp]{axessibility}  

\usepackage{hyperref}

\usepackage{orcidlink}

\begin{document}

\title{
BrushNet: A Plug-and-Play Image Inpainting Model with Decomposed Dual-Branch Diffusion
\vspace{-0.35cm}
} 
 
\titlerunning{BrushNet}

\author{
\small
Xuan Ju\inst{1,2} \and
Xian Liu\inst{1,2} \and
Xintao Wang\inst{1}\thanks{Corresponding author.} \and
Yuxuan Bian\inst{2} \and
Ying Shan\inst{1} \and
Qiang Xu\inst{2}$^\star$
}

\authorrunning{Xuan Ju, et al.}

\institute{
\vspace{-0.2cm}
$^1$ARC Lab, Tencent PCG  \xspace
$^2$The Chinese University of Hong Kong \\
\url{https://github.com/TencentARC/BrushNet}
}

\maketitle

\begin{center}
    \begin{figure}[htbp]
    \centering
    \vspace{-1.1cm}
    \includegraphics[width=0.95\linewidth]{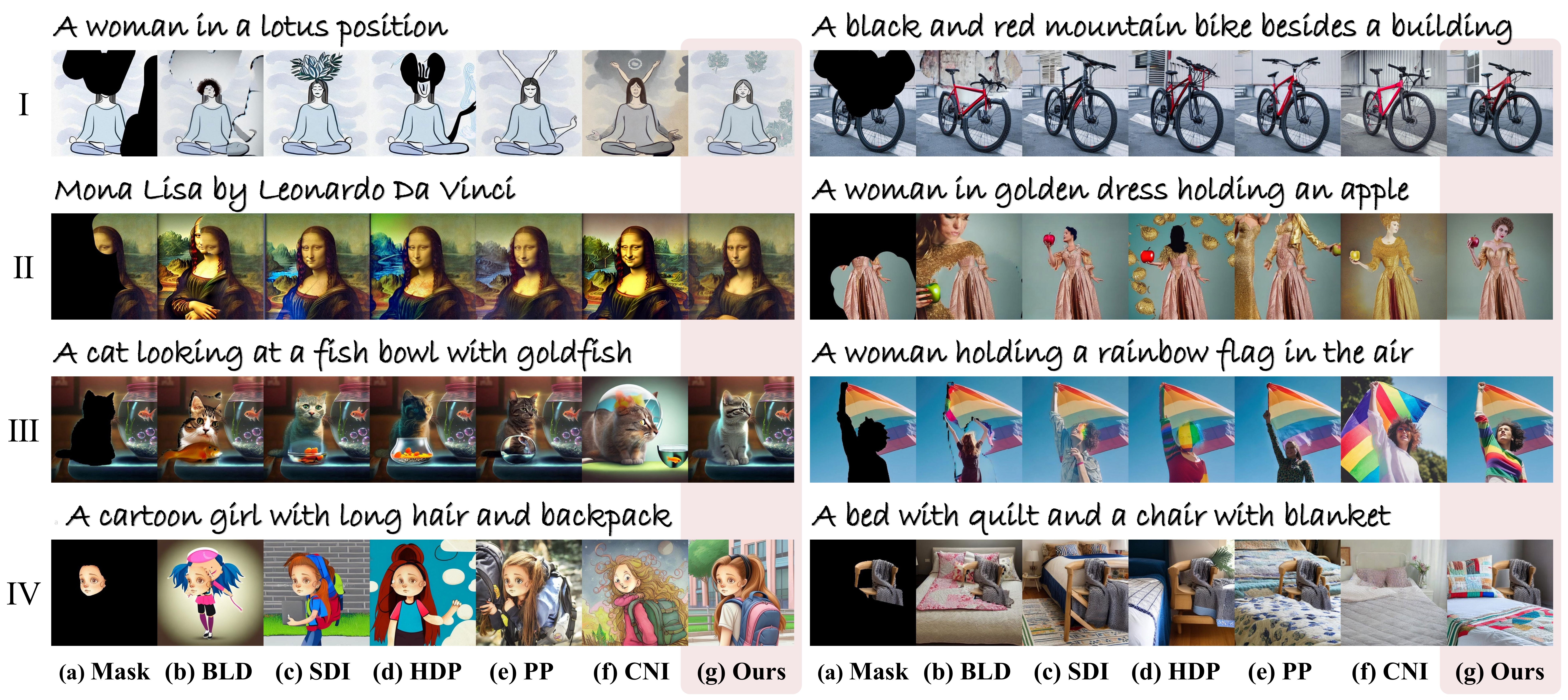}
    \vspace{-0.3cm}
    \caption{\textbf{Performance comparisons of \OurMethod~and previous image inpainting methods} across various inpainting tasks: (\uppercase\expandafter{\romannumeral1}) Random Mask (< 50\% masked), (\uppercase\expandafter{\romannumeral2}) Random Mask (> 50\% masked), (\uppercase\expandafter{\romannumeral3}) Segmentation Mask Inside-Inpainting, (\uppercase\expandafter{\romannumeral4}) Segmentation Mask Outside-Inpainting. Each group of results contains an artificial image (left) and a natural image (right) with $6$ inpainting methods: (b) Blended Latent Diffusion (BLD)~\cite{avrahami2023blended}, (c) Stable Diffusion Inpainting (SDI)~\cite{Rombach_2022_CVPR}, (d) HD-Painter (HDP)~\cite{manukyan2023hd}, (e) PowerPaint (PP)~\cite{zhuang2023task}, (f) ControlNet-Inpainting (CNI)~\cite{zhang2023adding}, and (g) Ours. %\OurMethod~ shows its superior coherence in (1) style, (2) content, (3) color, and (4) prompt alignment.
    }
    \label{fig:teaser}
    \vspace{-1.6cm}
\end{figure}

\end{center}

\begin{abstract} 
Image inpainting, the process of restoring corrupted images, has seen significant advancements with the advent of diffusion models (DMs). Despite these advancements, current DM adaptations for inpainting, which involve modifications to the sampling strategy or the development of inpainting-specific DMs, frequently suffer from semantic inconsistencies and reduced image quality.  
Addressing these challenges, our work introduces a novel paradigm: the division of masked image features and noisy latent into separate branches. This division dramatically diminishes the model's learning load, facilitating a nuanced incorporation of essential masked image information in a hierarchical fashion. 
Herein, we present \OurMethod, a novel plug-and-play dual-branch model engineered to embed pixel-level masked image features into any pre-trained DM, guaranteeing coherent and enhanced image inpainting outcomes. 
Additionally, we introduce \TrainingData~and \Benchmark~to facilitate segmentation-based inpainting training and performance assessment. Our extensive experimental analysis demonstrates BrushNet's superior performance over existing models across seven key metrics, including image quality, mask region preservation, and textual coherence.
\vspace{-0.5cm}

\end{abstract}

\section{Introduction}
\label{sec:introduction}

\vspace{-0.3cm}

Image inpainting~\cite{xu2023review} aims at restoring the missing regions of an image while maintaining the overall coherence. 
As a long-standing computer vision problem, it facilitates numerous applications such as virtual try-on~\cite{li2023virtual} and image editing~\cite{huang2024editingsurvey}.
Recently, diffusion models~\cite{ho2020denoising,song2020denoising} have demonstrated impressive performance in image generation, enabling flexible user control with semantic and structural conditions~\cite{Rombach_2022_CVPR, zhang2023adding}. To this end, researchers resort to diffusion-based pipelines for high-quality image inpainting that aligns with given text prompts.

Commonly used diffusion-based text-guided inpainting methods can be roughly divided into two categories: 
(1) \textit{Sampling strategy modification}~\cite{lugmayr2022repaint,avrahami2022blended,avrahami2023blended,liu2023image,zhang2023coherent,corneanu2024latentpaint,yang2023magicremover}, which modifies the standard denoising process by sampling the masked regions from a pre-trained diffusion model, and the unmasked areas are simply copy-pasted from the given image in each denoising step. 
Although they can be used in arbitrary diffusion backbones, the limited perceptual knowledge of mask boundaries and the unmasked image region context leads to incoherent inpainting results. 
(2) \textit{Dedicated inpainting models}~\cite{xie2023smartbrush,Rombach_2022_CVPR,zhuang2023task,xie2023dreaminpainter,wang2023imagen,ReplaceAnything,yu2023inpaint,yang2023uni}, which fine-tune a specially designed image inpainting model by expanding the input channel dimension of base diffusion models to incorporate provided corrupted image and mask.
While they enable the diffusion model to generate more satisfying results with specialized content-aware and shape-aware models, we argue, \textbf{is this architecture the best fit for diffusion-based inpainting?}

\begin{figure}[htbp]
\vspace{-0.6cm}
    \centering
    \includegraphics[width=0.95\linewidth]{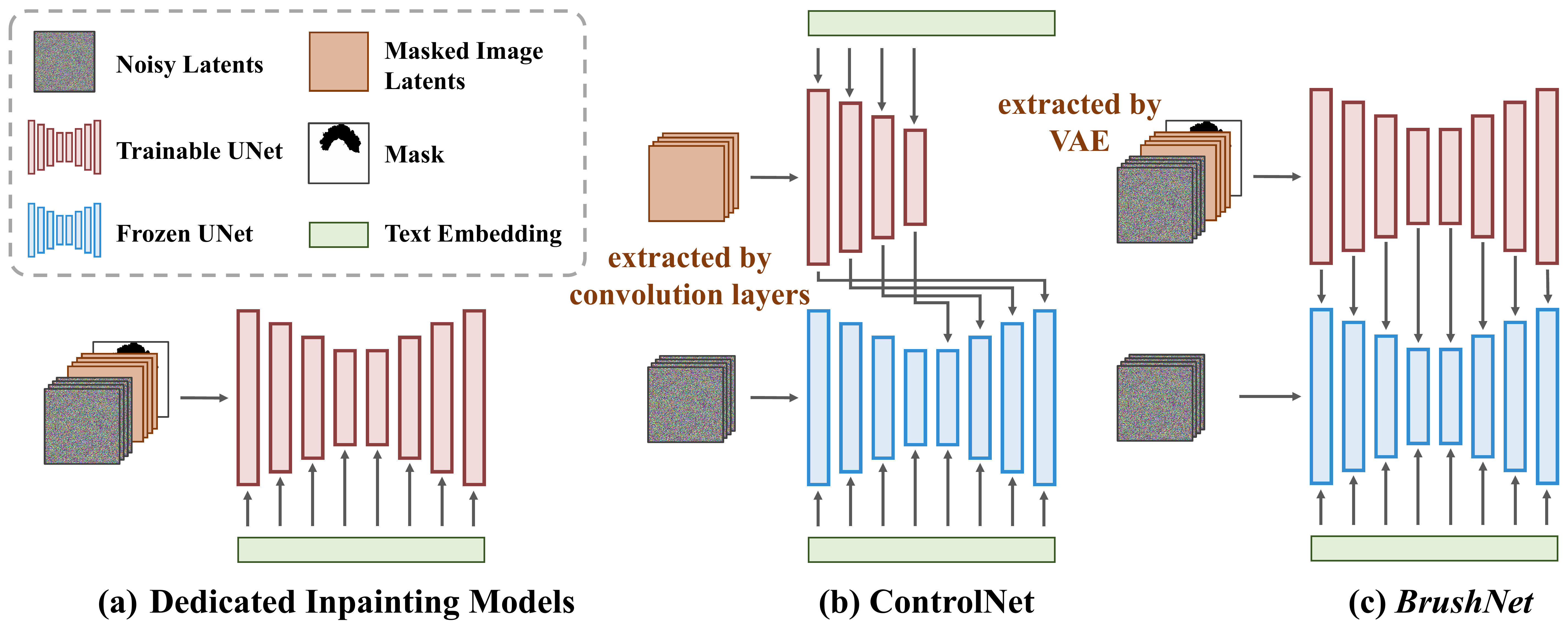}
\vspace{-0.2cm}
    \caption{\textbf{Comparison of previous inpainting architectures and \OurMethod.}}
    \label{fig:compare}
\vspace{-0.7cm}
\end{figure}

As shown in Fig.~\ref{fig:compare}, dedicated inpainting models fuse noisy latent, masked image latent, mask, and text at an early stage. 
This architectural design makes the masked image feature easily influenced by the text embedding, preventing subsequent layers in the UNet from obtaining pure masked image features due to the influence of text. 
Additionally, handling the condition and generation in a single branch imposes extra burdens on the UNet framework. 
These approaches also necessitate fine-tuning in different variations of diffusion backbones, which can be time-consuming with limited transferability.

Adding an additional branch dedicated to masked image feature extraction could be a promising approach to tackle the issue above. 
However, the existing solutions such as ControlNet~\cite{zhang2023adding} lead to inadequate information extraction and insertion when directly applied to inpainting, which originates from differences between controllable image generation and inpainting: inpainting requires pixel-to-pixel constraints with strong restrictive information instead of sparse structural control relying on text for content completion. As a result, ControlNet yields unsatisfactory results compared to dedicated inpainting models.

To address this issue, we propose \OurMethod, which introduces an additional branch to the diffusion framework, creating a more suitable architecture for image inpainting. Specifically, our designs are three-fold: 
(1) To improve the extraction of image features for adaptation to the UNet distribution, we use a VAE encoder instead of randomly initialized convolution layers to process the masked image. 
(2) To enable dense per-pixel control, we adopt a hierarchical approach by gradually incorporating the full UNet feature layer-by-layer into the pre-trained UNet. 
(3) To ensure pure image information is considered in the additional branch, we remove text cross-attention from UNet.
This design further offers plug-and-play capabilities and flexible unmasked region controllability to the inpainting process.
For better consistency and a larger range of unmasked region controllability, we additionally propose a blurred blending strategy.

To ensure a comprehensive evaluation for real-life applications, we categorize inpainting tasks into two distinct types based on mask shape: random brush masks and segmentation-based masks. 
We utilize EditBench~\cite{wang2023imagen} as the comparing benchmark for random brush mask inpainting. 
Additionally, we introduce a new training dataset \TrainingData~and a new benchmark \Benchmark~for training and evaluating segmentation-based mask inpainting. 
Results show \OurMethod~achieve state-of-the-art performance across $7$ metrics encompassing image quality, masked region preservation, and text alignment.

\vspace{-0.25cm}    

\section{Related Work}
\label{sec:related_work}

\vspace{-0.3cm}    

Image inpainting is a classic problem in computer vision, aiming to restore masked regions of an image with plausible and natural content~\cite{quan2024deep,xu2023review}. 
Previous methods based on traditional techniques~\cite{bertalmio2000image,criminisi2004region}, Variational Auto-Encoders (VAEs)~\cite{zheng2019pluralistic,peng2021generating} and Generative Adversarial Networks (GANs)~\cite{liu2021pd,zhao2021large,zheng2022image} often require auxiliary hand-engineered features but yield poor results.
Recently, diffusion-based methods~\cite{lugmayr2022repaint,avrahami2022blended,avrahami2023blended,xie2023smartbrush,Kandinsky_2,liu2023image} gain popularity due to their impressive high-quality generation, fine-grained control, and output diversity~\cite{ho2020denoising,ho2022classifier,Rombach_2022_CVPR}.

Initial attempts at utilizing diffusion models for text-guided inpainting~\cite{lugmayr2022repaint,avrahami2022blended,avrahami2023blended,liu2023image,zhang2023coherent,corneanu2024latentpaint,yang2023magicremover}, such as Blended Latent Diffusion, modifies the standard denoising strategy by sampling the masked regions from a pre-trained diffusion model and the unmasked areas from the given image, which is commonly used as the default inpainting choice in widely-used image generation libraries like Diffusers~\cite{von-platen-etal-2022-diffusers}.
Although these methods demonstrate satisfactory results in simple image inpainting tasks and can be plug-and-play to any diffusion model, they struggle with complex mask shapes, image contents, and text prompts, leading to results that lack coherence.
This is primarily attributed to their limited perceptual knowledge of mask boundaries and the unmasked image region context.

Previous works~\cite{xie2023smartbrush,Rombach_2022_CVPR,zhuang2023task,xie2023dreaminpainter,wang2023imagen,ReplaceAnything,yu2023inpaint,yang2023uni} address this issue by fine-tuning the base models into content-aware and shape-aware models specifically designed for image inpainting.
Specifically, SmartBrush~\cite{xie2023smartbrush} augments the diffusion U-Net with object-mask prediction to guide the sampling process with mask boundaries information.
Stable Diffusion Inpainting~\cite{Rombach_2022_CVPR} fine-tunes a diffusion model specifically designed for inpainting tasks, taking the mask, masked image, and noisy latent as inputs to UNet architecture.
HD-Painter~\cite{manukyan2023hd} and PowerPaint~\cite{zhuang2023task} build upon the foundation of Stable Diffusion Inpainting, separately enhancing the generation quality and enabling the model to perform multiple tasks. 

However, these approaches make it difficult to effectively transfer their inpainting ability to arbitrary pre-trained models, restricting their applicability. 
To enable any diffusion model with inpainting capabilities, the community fine-tunes ControlNet~\cite{zhang2023adding} on inpainting image pairs. 
However, the model design of ControlNet exhibits limitations in its perceptual understanding of masks and masked images, which consequently leads to unsatisfactory outcomes.
Compared with previous methods (shown in Tab.~\ref{tab:compare_other_methods}), \OurMethod~is plug-and-play, content-aware, and shape-aware, with a flexible preserving degree for unmasked regions.

\begin{table}[htbp]
\vspace{-0.5cm}
    \centering
    \small
    \caption{\textbf{Comparison of \OurMethod~with Previous Image Inpainting Methods.} \OurMethod~offers the advantage of being plug-and-play with any pretrained diffusion model. Moreover, it allows for flexible control over the scale of inpainting and is designed to be aware of both the mask shape and the unmasked content. Note that we only list commonly used text-guided diffusion methods in this table. }
\vspace{-0.25cm}
    \scalebox{0.93}{
    \renewcommand\arraystretch{0.8}
\setlength{\tabcolsep}{0.9mm}{
    \begin{tabular}{ccccc}
\toprule
Model                       & Plug-and-Play  & Flexible-Scale & Content-Aware & Shape-Aware \\ \midrule
Blended Diffusion~\cite{avrahami2022blended, avrahami2023blended}           &  \checkmark    &       &        &       \\
SmartBrush~\cite{xie2023smartbrush}           &      &       &         &  \checkmark  \\
SD Inpainting~\cite{Rombach_2022_CVPR} &               &      &  \checkmark  & \checkmark \\
PowerPaint~\cite{zhuang2023task}           &      &       &   \checkmark     &   \checkmark    \\
HD-Painter~\cite{manukyan2023hd}           &      &       &   \checkmark     &   \checkmark    \\
ReplaceAnything~\cite{ReplaceAnything}             &               &      &  \checkmark  &  \checkmark \\ 
Imagen~\cite{wang2023imagen}             &               &      &  \checkmark  &  \checkmark \\ 
ControlNet-Inpainting~\cite{zhang2023adding}             &       \checkmark        &  \checkmark    &   \checkmark   &  \\ \midrule
\OurMethod            &   \checkmark  &    \checkmark &  \checkmark  &  \checkmark \\ \bottomrule
\end{tabular}}}
\vspace{-0.2cm}
    \label{tab:compare_other_methods}
\end{table}

\vspace{-0.9cm}
\section{Preliminaries and Motivation}
\label{sec:preliminaries_and_motivation}

\vspace{-0.3cm}

In this section, we will first introduce diffusion models in Sec.~\ref{sec:diffusion_models}. Then, Sec.~\ref{sec:previous_inpainting_models} would review previous inpainting techniques based on sampling strategy modification and special training. Finally, the motivation is outlined in Section~\ref{sec:motivation}.

\vspace{-0.4cm}

\subsection{Diffusion Models}
\label{sec:diffusion_models}

\vspace{-0.1cm}

Diffusion models include a forward process that adds Gaussian noise $\epsilon$ to convert clean sample $z_0$ to noise sample $z_T$, 
and a backward process that iteratively performs denoising from $z_T$ to $z_0$, where $\epsilon \sim \mathcal{N} \left( 0,1 \right)$, and $T$ represents the total number of timesteps. The forward process can be formulated as:

\begin{equation}
 z_t=\sqrt{\alpha _t}z_{0}+\sqrt{1-\alpha _t}\epsilon
 \label{eq:forward}
\end{equation}

$z_t$ is the noised feature at step $t$ with $t\sim \left[ 1,T \right]$, and $\alpha$ is a hyper-parameter.

In the backward process, given input noise $z_T$ sampled from a random Gaussian distribution, learnable network $\epsilon_{\theta}$ estimates noise at each step $t$ conditioned on $C$. After $T$ progressively refining iterations, $z_{0}$ is derived as the output sample:

\begin{equation}
z_{t-1}=\frac{\sqrt{\alpha _{t-1}}}{\sqrt{\alpha _t}}z_t+\sqrt{\alpha _{t-1}}\left( \sqrt{\frac{1}{\alpha _{t-1}}-1}-\sqrt{\frac{1}{\alpha _t}-1} \right) \epsilon _{\theta}\left( z_t,t,C \right) 
 \label{eq:ddim_sample}
\end{equation}

\vspace{-0.2cm}

The training of diffusion models revolves around optimizing the denoiser network $\epsilon_{\theta}$ to conduct denoising with condition $C$, guided by the objective:

\vspace{-0.2cm}

\begin{equation}
\underset{\theta}{\min}E_{z_0,\epsilon \sim \mathcal{N} \left( 0,I \right) ,t\sim U\left( 1,T \right)}\left\| \epsilon -\epsilon _{\theta}\left(z_t, t, C\right) \right\| 
 \label{eq:train_objective}
\end{equation}

\vspace{-0.5cm}

\subsection{Previous Inpainting Models} 
\label{sec:previous_inpainting_models}

\vspace{-0.1cm}

\paragraph{Sampling Strategy Modification.} 
This line of research achieves inpainting by gradually blending masked images with the generated results. 
The most used method among them is Blended Latent Diffusion (BLD)~\cite{avrahami2023blended}, serving as the default choice for inpainting in widely-used diffusion-based image generation libraries (\textit{e.g.}, Diffusers~\cite{von-platen-etal-2022-diffusers}). 
Given a binary mask $m$ and a masked image $x_0^{masked}$, BLD first extracts the latent representation $z_0^{masked}$ of the masked image using VAE. 
Subsequently, the mask $m$ is resized to $m^{resized}$ to match the size of the latent representation. 
To formulate the inpainting process, BLD adds Gaussian noise to $z_0^{masked}$ for $T$ steps and gets $z_t^{masked}$, where $t\sim \left[ 1,T \right]$. 
Then, denoising steps start from $z_T^{masked}$, where each sampling step in eq.~\ref{eq:ddim_sample} is followed by:

\vspace{-0.1cm}

\begin{equation}
z_{t-1} \gets z_{t-1}\cdot\left(1-m^{resized}\right)+z_{t-1}^{masked}\cdot m^{resized}
 \label{eq:blending}
\end{equation}

Despite its simplicity in implementation, BLD exhibits suboptimal performance in terms of both unmasked region preservation and generation content alignment. 
This is due to (1) the resize of the mask preventing it from correctly blending the noisy latent, (2) the diffusion model lacking perceptual knowledge of mask boundaries and the unmasked image region context.

\vspace{-0.2cm}

\paragraph{Dedicated Inpainting Models.}

To enhance the performance of inpainting, previous works fine-tune the base models by expanding the input UNet channel to include the mask and masked image inputs, turning it into an architecture specifically designed for image inpainting. 
Though having better generation results compared with BLD, they still have several drawbacks: 
(1) These models merge the noisy latent, masked image latent, and mask at the initial convolution layer of the UNet architecture, where they are collectively influenced by the text embedding.
Consequently, subsequent layers in the UNet model struggle to obtain pure masked image features due to the text's influence.
(2) Incorporating both the condition processing and generation within a single branch places additional burdens on the UNet framework.
(3) These approaches require extensive fine-tuning across various variations of diffusion backbones, which is computationally intensive and lacks transferability to custom diffusion models.

\vspace{-0.3cm}

\subsection{Motivation}
\label{sec:motivation}

\vspace{-0.15cm}

Based on the analysis presented in Section~\ref{sec:previous_inpainting_models}, a more effective architecture design of inpainting would be introducing an additional branch specifically dedicated to masked image processing.
ControlNet~\cite{zhang2023adding} is one of the widely adopted strategies that exemplifies this idea.
However, it should be noted that directly fine-tuning ControlNet, which is originally designed for controllable image generation, on the inpainting task yields unsatisfactory results.
ControlNet designs a lightweight encoder to incorporate out-of-domain structural conditions (\textit{e.g.}, skeleton) and relies on text guidance for content generation, which is unsuitable for pixel-level inpainting image feature injection.
Furthermore, ControlNet typically relies on sparse control, meaning that merely adding control to residuals in the UNet framework would be sufficient, while inpainting requires pixel-to-pixel constraints with strong restrictive information. 
Thus a new architecture specifically designed for inpainting is urgently needed.

\vspace{-0.35cm}

\section{Method}
\label{sec:method}

\vspace{-0.2cm}

\begin{figure}[htbp]
    \centering
    \includegraphics[width=1.\linewidth]{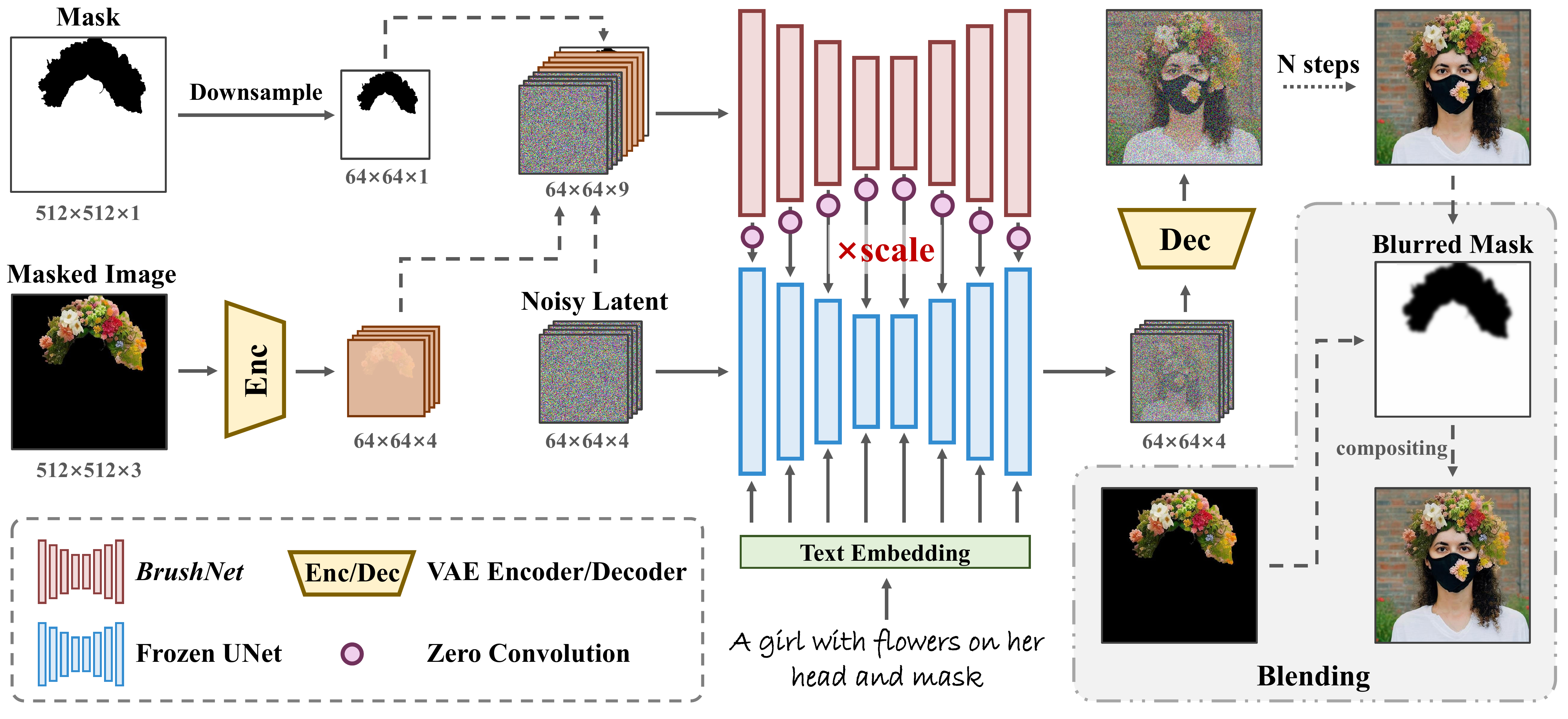}
    \vspace{-0.1cm}
    \caption{\textbf{Model overview.} Our model outputs an inpainted image given the mask and masked image input. Firstly, we downsample the mask to accommodate the size of the latent, and input the masked image to the VAE encoder to align the distribution of latent space. Then, noisy latent, masked image latent, and downsampled mask are concatenated as the input of \OurMethod. The feature extracted from \OurMethod~is added to pretrained UNet layer by layer after a zero convolution block~\cite{zhang2023adding}. After denoising, the generated image and masked image are blended with a blurred mask.}
    \label{fig:model}
\vspace{-1.1cm}
\end{figure}

An overview of \OurMethod~is shown in Fig.~\ref{fig:model}. We employ a dual-branch strategy for masked image guidance insertion (Sec.~\ref{sec:masked_image_guidance}). Blending operations with a blurred mask is used to ensure better-unmasked region preservation (Sec.~\ref{sec:blending_operation}). Notably, \OurMethod~can achieve flexible control by adjusting the added scale.

\vspace{-0.25cm}

\subsection{Masked Image Guidance}
\label{sec:masked_image_guidance}

\vspace{-0.15cm}

The insertion of the masked image feature into the pre-trained diffusion network is accomplished through an additional branch, which explicitly separates the feature extraction of masked images from the image-generating processes.
The input to the additional branch includes noisy latent, masked image latent, and downsampled mask, which are concatenated together to form the input. 
Specifically, the noisy latent provides generation information during the present generation process, helping \OurMethod~enhance the semantic coherence of the masked image feature.
The masked image latent is extracted from the masked image using VAE, which is aligned with the data distribution of the pre-trained UNet.
To ensure the alignment of the mask size with the noisy latent and masked image latent, we employ cubic interpolation to downsample the mask.

To process the masked image features, \OurMethod~utilizes a clone of the pre-trained diffusion model while excluding its cross-attention layers. The pre-trained weights of the diffusion model serve as a strong prior for extracting the masked image features, while the removal of the cross-attention layers ensures that only pure image information is considered within this additional branch.
\OurMethod~feature is inserted into the frozen diffusion model layer-by-layer, enabling dense per-pixel control hierarchically. 
Similar to ControlNet~\cite{zhang2023adding}, we employ zero convolution layers to establish a connection between the locked model and the trainable \OurMethod. 
This ensures that harmful noise does not influence the hidden states in the trainable copy during the initial stages of training.

The feature insertion operation is shown in Eq.~\ref{eq:insertion}. Specifically, $\epsilon _{\theta}\left( z_t,t,C \right) _i$ indicates the feature of the $i$-th layer in network $\epsilon _{\theta}$ with $i\sim \left[ 1,n \right]$, where $n$ is the number of layers. The same notation applies to $\epsilon _{\theta}^{BrushNet}$. $\epsilon _{\theta}^{BrushNet}$ takes the concatenated noisy latent $z_t$, masked image latent $z_{0}^{masked}$, and downsampled mask $m^{resized}$ as input, with the concatenation operation denoted as $\left[ \cdot \right]$. $\mathcal{Z}$ is the zero convolution operation. $w$ is the preservation scale used to adjust the influence of \OurMethod~on pretrained diffusion model. 

\vspace{-0.4cm}

\begin{equation}
\epsilon _{\theta}\left( z_t,t,C \right) _i=\epsilon _{\theta}\left( z_t,t,C \right) _i+w\cdot \mathcal{Z} \left( \epsilon _{\theta}^{BrushNet}\left( \left[ z_t,z_{0}^{masked},m^{resized} \right] ,t \right) _i \right) 
 \label{eq:insertion}
\end{equation}

\vspace{-0.5cm}

\subsection{Blending Operation}
\label{sec:blending_operation}

\vspace{-0.1cm}

As mentioned in Section~\ref{sec:blending_operation}, the blending operation conducted in latent space can result in inaccuracies due to the resizing of the mask. 
Similarly, in our approach, a similar issue arises as we resize the mask to match the size of the latent space, which can introduce potential inaccuracies.
Additionally, it is important to acknowledge that VAE encoding and decoding operations have inherent limitations and may not ensure complete image reconstruction.

To ensure a fully consistent image reconstruction of the unmasked region, previous works have explored different techniques. 
Some approaches~\cite{zhuang2023task,ReplaceAnything}, utilize past-and-copy methods, where the unmasked region is directly copied from the original image. 
However, this can result in a lack of semantic coherence in the final generation results.
On the other hand, adopting latent blending operations inspired by BLD~\cite{avrahami2023blended,Rombach_2022_CVPR} has been observed to face challenges in effectively preserving the desired information in the unmasked regions.

In this work, we present a simple pixel space solution to address this issue by first blurring the mask and then performing copy-and-paste using the blurred mask. Although this approach may result in a slight loss of accuracy in preserving the details of the mask boundary, the error is nearly imperceptible to the naked eye and results in significantly improved coherence in the mask boundary.

\subsection{Flexible Control}
\label{sec:flexible_control}

\vspace{-0.1cm}

The architecture design of \OurMethod~inherently makes it suitable for seamless plug-and-play integration to various pretrained diffusion models and enables flexible preservation scale. 
Specifically, the flexible control of our proposed \OurMethod~includes: 
(1) Since \OurMethod~does not modify the weights of the pretrained diffusion model, it can be readily integrated as a plug-and-play component with any community fine-tuned diffusion models. This allows for easy adoption and experimentation with different pretrained models.
(2) Preservation Scale Adjustment: The preservation scale of the unmasked region can be controlled by incorporating \OurMethod~features into the frozen diffusion model with the weight $w$. This weight determines the influence of \OurMethod~on the preservation scale, offering the ability to adjust the desired level of preservation.
(3) Blurring Scale and Blending Operation: By adjusting the scale of blurring and deciding whether to apply the blending operation, the preservation scale of the unmasked region can be further customized.
These features allow for flexible fine-grained control over the inpainting process. More explanation can be found in Sec.~\ref{sec:flexible_control_ability}.

\vspace{-0.35cm}

\section{Experiments}
\label{sec:experiments}

\vspace{-0.25cm}

\subsection{Evaluation Benchmark and Metrics}

\vspace{-0.15cm}

\paragraph{\textbf{Benchmark.}}
Previous commonly used datasets in the image inpainting field include CelebA~\cite{liu2015faceattributes}, CelebA-HQ~\cite{huang2018introvae}, ImageNet~\cite{deng2009imagenet}, MSCOCO~\cite{lin2014microsoft}, Open Images~\cite{kuznetsova2020open}, and LSUN-Bedroom~\cite{yu2015lsun}. 
However, these datasets either primarily focus on a small area, such as human faces, or predominantly consist of low-quality, cluttered real-life scene data. 
As a result, these datasets are not well-suited for training and evaluating diffusion-based inpainting models, which can generate high-quality diverse images that align with text prompts.

Recently proposed EditBench~\cite{wang2023imagen} serves as a benchmark specifically designed for text-guided image inpainting for diffusion models. 
This benchmark consists of a collection of $240$ images comprising an equal ratio of natural images and generated images, with mask and caption annotation for each image. 
However, the annotated masks in EditBench are mostly random shapes without specific object information, neglecting the practical application of inpainting in real scenarios such as replacing an object with an external mask, as commonly seen in E-commerce product displays and image editing.

\begin{figure}[htbp]
    \centering
    \includegraphics[width=0.99\linewidth]{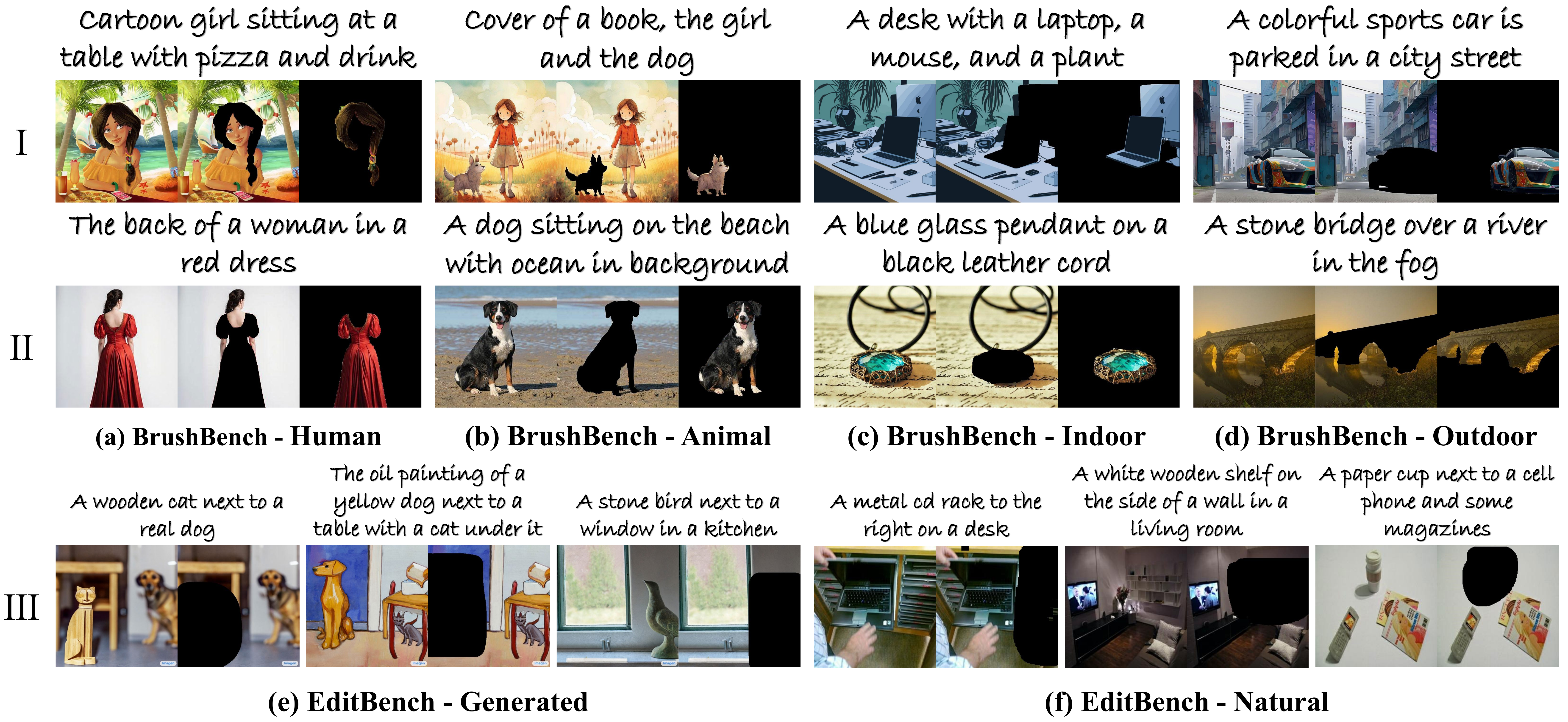}
    \vspace{-0.2cm}
    \caption{\textbf{Benchmark overview.} \uppercase\expandafter{\romannumeral1} and \uppercase\expandafter{\romannumeral2} separately show natural and artificial images, masks, and caption of \Benchmark. (a) to (d) show images of humans, animals, indoor scenarios, and outdoor scenarios. Each group of images shows the original image, inside-inpainting mask, and outside-inpainting mask, with an image caption on the top. \uppercase\expandafter{\romannumeral3} show image, mask, and caption from EditBench~\cite{wang2023imagen}, with (e) for generated images and (f) for natural images. The images are randomly selected from both benchmarks.}
    \label{fig:benchmark}
    \vspace{-0.7cm}
\end{figure}

To fill the gap, we propose \Benchmark~for segmentation-based inpainting, as shown in Fig.~\ref{fig:benchmark}.
\Benchmark~comprises a total of $600$ images, with each image accompanied by the human-annotated mask and caption annotation. 
The images in \Benchmark~are evenly distributed between natural images and artificial images, such as paintings. 
Furthermore, the dataset ensures an equal distribution among different categories, including humans, animals, indoor scenarios, and outdoor scenarios. 
This balanced distribution enables a fair evaluation across various categories, promoting better evaluation equity.

To further enhance the analysis of the inpainting task, we categorize it into two distinct types based on the masks used: random brush masks and segmentation-based masks. 
We use EditBench as the comparison benchmark for random brush masks and use \Benchmark~for segmentation-based masks. 
Regarding inpainting with segmentation-based masks, we refine the task by considering two specific scenarios: segmentation mask inside-inpainting and segmentation mask outside-inpainting. By separating these two subtasks, we can better understand the inpainting performance in different image regions.

\vspace{-0.25cm}

\paragraph{\textbf{Dataset.}} To train segmentation-based mask inpainting, we annotate segmentation mask on Laion-Aesthetic~\cite{schuhmann2022laion} dataset, called \TrainingData. We employ the Grounded-SAM~\cite{ren2024grounded} to annotate open-world masks and then filter the masks based on their confidence score, ensuring that only masks with relatively higher confidence scores are retained. Additionally, we consider factors such as reasonable mask size and good mask continuity during the filtering process. \footnote{The proposed \TrainingData~and \Benchmark~will be released along with the codes.}

\vspace{-0.25cm}

\paragraph{\textbf{Metrics.}} We consider $7$ metrics from three aspects: image generation quality, masked region preservation, and text alignment.

\vspace{-0.2cm}

\begin{itemize}
    \item[$\bullet$] \textit{Image Generation Quality.} 
    Metrics most used by previous inpainting methods (\textit{e.g.}, FID~\cite{fid} and KID~\cite{kid}) show a poor representation of the rich and varied content generated by modern text-to-image models~\cite{jayasumana2023rethinking}. 
    Thus, we use Image Reward (\textbf{IR})~\cite{xu2023imagereward}, HPS v2 (\textbf{HPS})~\cite{wu2023human}, and Aesthetic Score (\textbf{AS})~\cite{schuhmann2022laion} that align with human perception. 
    Specifically, ImageReward and HPS v2 are text-to-image human preference evaluation models trained on large-scale datasets of human preference choices on generated images. Aesthetic Score is a linear model trained on image quality rating pairs of real images.
    
    \item[$\bullet$] \textit{Masked Region Preservation.} We follow previous works using standard Peak Signal-to-Noise Ratio (\textbf{PSNR})~\cite{psnr}, Learned Perceptual Image Patch Similarity (\textbf{LPIPS})~\cite{zhang2018unreasonable}, and Mean Squared Error (\textbf{MSE})~\cite{mse} in the unmasked region among the generated image and the original image.
    \item[$\bullet$] \textit{Text Alignment.} We use CLIP Similarity (\textbf{CLIP Sim}) ~\cite{clipsim} to evaluate text-image consistency between the generated images and corresponding text prompts. CLIP Similarity projects text and images to the same shared space with the CLIP model~\cite{clip} and evaluates the similarity of their embeddings.
\end{itemize}

\vspace{-0.55cm}

\subsection{Implementation Details}

\vspace{-0.1cm}

We perform the inference of different inpainting methods in the same setting unless specifically clarified, \emph{i.e.}, on NVIDIA Tesla V100 following their open-source code with a base model of Stabe Diffusion v1.5 in 50 steps, with a guidance scale of 7.5. 
We keep the recommended hyper-parameter for each inpainting method in all images for fair comparison. 
\OurMethod~and all ablation models are trained for $430$ thousands steps on 8 NVIDIA Tesla V100 GPUs, which takes around 3 days.
For comparison on \Benchmark, we use \OurMethod~trained on \TrainingData. For comparison on EditBench, we use the model trained on LAION-5B with random masks.
Details can be found in the provided codes. 

\vspace{-0.35cm}

\subsection{Quantitative Comparison}

\vspace{-0.8cm}

\begin{table}[htbp]
\centering
\scriptsize
\caption{\textbf{Quantitative comparisons among \OurMethod~and other diffusion-based inpainting models in \Benchmark}: Blended Latent Diffusion (BLD)~\cite{avrahami2023blended}, Stable Diffusion Inpainting (SDI)~\cite{Rombach_2022_CVPR}, HD-Painter (HDP)~\cite{manukyan2023hd}, PowerPaint (PP)~\cite{zhuang2023task}, and ControlNet-Inpainting (CNI)~\cite{zhang2023adding}. Metrics encompassing image quality, masked region preservation, and text alignment (Text Align) for inside-inpainting and outside-inpainting are shown in the table. All models use Stable Diffusion V1.5 as base model. \textcolor{Red}{\textbf{Red}} stands for the best result, \textcolor{Blue}{\textbf{Blue}} stands for the second best result. }
\vspace{-0.2cm}
\scalebox{1.}{
\setlength{\tabcolsep}{0.9mm}{
\begin{threeparttable}
{
\begin{tabular}{cl|ccc|ccc|c}
\toprule
\toprule
\multicolumn{2}{c|}{\textbf{Metrics}}  & \multicolumn{3}{c|}{\textbf{Image Quality}} & \multicolumn{3}{c|}{\textbf{Masked Region Preservation}} & \textbf{Text Align} \\
\midrule
\multicolumn{2}{c|}{\textbf{Models}}    & \textbf{IR}$_{^{\times 10}}$$\uparrow$   & \textbf{HPS}$_{^{\times 10^2}}$$\uparrow$  & \textbf{AS}$\uparrow$   & \textbf{PSNR}$\uparrow$     & \textbf{LPIPS}$_{^{\times 10^3}}$$\downarrow$  & \textbf{MSE}$_{^{\times 10^3}}$$\downarrow$      & \textbf{CLIP Sim}$\uparrow$     \\ \midrule
\multirow{7}{*}{\rotatebox{90}{\textbf{Inside}}}  & \textbf{\xspace BLD}~\cite{avrahami2023blended} & 9.78&25.87&6.17&21.33&9.76&49.26&26.15 \\
& \textbf{\xspace SDI~\cite{Rombach_2022_CVPR}} &  11.72&27.06&6.50&21.52&13.87&48.39&26.17 \\
& \textbf{\xspace HDP}~\cite{manukyan2023hd} &11.68&26.90&6.42&22.61&9.95&43.50&26.37  \\
& \textbf{\xspace PP}~\cite{zhuang2023task} &  11.46&27.35&6.24&21.43&32.73&48.43&\textcolor{Red}{\textbf{26.48}} \\
& \textbf{\xspace CNI}~\cite{zhang2023adding} & 9.9&26.02&\textcolor{Red}{\textbf{6.53}}&12.39&78.78&243.62&\textcolor{Blue}{\textbf{26.47}}   \\ 
& \textbf{\xspace CNI}*~\cite{zhang2023adding} &  11.21&26.92&6.39&\textcolor{Blue}{\textbf{22.73}}&24.58&\textcolor{Blue}{\textbf{43.49}}&26.22  \\ 
& \textbf{\xspace Ours} & \textcolor{Blue}{\textbf{12.36}}&\textcolor{Blue}{\textbf{27.40}}&\textcolor{Red}{\textbf{6.53}}&21.65&\textcolor{Blue}{\textbf{9.31}}&48.28&\textcolor{Red}{\textbf{26.48}}    \\ 
& \textbf{\xspace Ours}* & \textcolor{Red}{\textbf{12.64}}&\textcolor{Red}{\textbf{27.78}}&\textcolor{Blue}{\textbf{6.51}}&\textcolor{Red}{\textbf{31.94}}&\textcolor{Red}{\textbf{0.80}}&\textcolor{Red}{\textbf{18.67}}&26.39   \\ 
\midrule
\multirow{7}{*}{\rotatebox{90}{\textbf{Outside}}} &  \textbf{\xspace BLD}~\cite{avrahami2023blended} &7.81&26.77&6.23&15.85&35.86&21.40&26.73\\
&  \textbf{\xspace SDI}~\cite{Rombach_2022_CVPR} &  10.27&27.99&\textcolor{Blue}{\textbf{6.55}}&18.04&19.87&15.13&27.21  \\
& \textbf{\xspace HDP}~\cite{manukyan2023hd} &    9.66&27.79&6.46&18.03&22.99&15.22&26.96 \\
& \textbf{\xspace PP~\cite{zhuang2023task}} &   7.45&28.01&6.26&18.04&31.78&15.13&26.72 \\ 
& \textbf{\xspace CNI}~\cite{zhang2023adding} &  9.26&27.68&6.42&11.91&83.03&58.16&\textcolor{Blue}{\textbf{27.29}}  \\ 
& \textbf{\xspace CNI}*~\cite{zhang2023adding} &  9.57&27.76&6.28&17.50&37.72&19.95&26.92  \\ 
& \textbf{\xspace Ours} &  \textcolor{Blue}{\textbf{10.82}}&\textcolor{Blue}{\textbf{28.02}}&\textcolor{Red}{\textbf{6.64}}&\textcolor{Blue}{\textbf{18.06}}&\textcolor{Blue}{\textbf{22.86}}&\textcolor{Blue}{\textbf{15.08}}&\textcolor{Red}{\textbf{27.33}} \\ 
& \textbf{\xspace Ours}* & \textcolor{Red}{\textbf{10.88}}&\textcolor{Red}{\textbf{28.09}}&\textcolor{Red}{\textbf{6.64}}&\textcolor{Red}{\textbf{27.82}}&\textcolor{Red}{\textbf{2.25}}&\textcolor{Red}{\textbf{4.63}}&27.22 \\ 
     \bottomrule
     \bottomrule
\end{tabular}
\begin{tablenotes}
\footnotesize
\item[*] with blending operation
\end{tablenotes}
}
\end{threeparttable}
}
}
\vspace{-0.7cm}
\label{tab:ours_bench}
\end{table}

Tab.~\ref{tab:ours_bench} and Tab.~\ref{tab:editbench} show the quantitative comparison on \Benchmark~and EditBench~\cite{wang2023imagen}. 
We compare the inpainting results of sampling strategy modification method Blended Latent Diffusion~\cite{avrahami2023blended}, dedicated inpainting models Stable Diffusion Inpainting~\cite{Rombach_2022_CVPR}, HD-Painter~\cite{manukyan2023hd}, and PowerPaint~\cite{zhuang2023task}, as well as plug-and-play method ControlNet~\cite{zhang2023adding} trained on inpainting data.

Results demonstrate \OurMethod's effectiveness across image quality, masked region preservation, and image-text alignment. 
Blended Latent Diffusion~\cite{avrahami2023blended} shows the poorest results in image quality and text alignment, derived from the incoherence between the generated masked and the unmasked given image. 
At the same time, its performance in masked region preservation is also not satisfactory because of the loss incurred by resized mask blending operation in the latent space. 
Modified from Stable Diffusion Inpainting~\cite{Rombach_2022_CVPR}, HD-Painter~\cite{manukyan2023hd} and PowerPaint~\cite{zhuang2023task}demonstrate comparable performance to Stable Diffusion Inpainting in the task of inside-inpainting.
However, when it comes to outside-inpainting, their results in terms of image quality and text alignment are significantly poorer compared to Stable Diffusion Inpainting, which can be attributed to their exclusive emphasis on the inside-inpainting task.

ControlNet~\cite{zhang2023adding} trained on inpainting has the most similar experimental configuration to ours.
Due to the mismatch between its model design and the inpainting task, ControlNet shows poor results in masked region preservation and image quality, necessitating its combination with Blended Latent Diffusion~\cite{avrahami2023blended} to generate satisfying inpainted images. 
However, even with this combination, it still falls short compared to dedicated inpainting models and \OurMethod.

\begin{table}[htbp]

\vspace{-0.5cm}

\centering
\scriptsize
\caption{\textbf{Quantitative comparisons among \OurMethod~and other diffusion-based inpainting models in EditBench}. A detailed explanation of compared methods and metrics can be found in the caption of Tab.~\ref{tab:ours_bench}. \textcolor{Red}{\textbf{Red}} stands for the best result, \textcolor{Blue}{\textbf{Blue}} stands for the second best result.}

\vspace{-0.3cm}

\scalebox{1.}{
\renewcommand\arraystretch{1.2}
\setlength{\tabcolsep}{0.9mm}{
\begin{threeparttable}
{
\begin{tabular}{l|ccc|ccc|c}
\toprule
\toprule
\textbf{Metrics}  & \multicolumn{3}{c|}{\textbf{Image Quality}} & \multicolumn{3}{c|}{\textbf{Masked Region Preservation}} & \textbf{Text Align} \\
\midrule
\textbf{Models}    & \textbf{IR}$_{^{\times 10}}$$\uparrow$   & \textbf{HPS}$_{^{\times 10^2}}$$\uparrow$  & \textbf{AS}$\uparrow$   & \textbf{PSNR}$\uparrow$     & \textbf{LPIPS}$_{^{\times 10^3}}$$\downarrow$  & \textbf{MSE}$_{^{\times 10^3}}$$\downarrow$      & \textbf{CLIP Sim}$\uparrow$     \\ \midrule
\textbf{BLD}~\cite{avrahami2023blended} & 0.90&23.81&5.44&20.89&10.93&31.90&28.62
\\
\textbf{SDI}~\cite{Rombach_2022_CVPR} & 1.86&24.24&5.69&23.25&6.94&24.30&28.00
 \\
\textbf{HDP}~\cite{manukyan2023hd} & 1.74&24.20&5.64&23.07&\textcolor{Blue}{\textbf{6.70}}&24.32&28.34
\\
\textbf{PP}~\cite{zhuang2023task} & 1.24&24.50&5.44&23.34&20.12&24.12&27.80
\\
\textbf{CNI}~\cite{zhang2023adding} &  1.49&24.46&\textcolor{Blue}{\textbf{5.82}}&12.71&69.42&159.71&28.16  \\ 
\textbf{CNI}*~\cite{zhang2023adding} & 0.90&23.79&5.46&22.61&35.93&26.14&27.74   \\ 
\textbf{Ours} & \textcolor{Blue}{\textbf{4.40}}&\textcolor{Blue}{\textbf{25.10}}&\textcolor{Red}{\textbf{5.84}}&\textcolor{Blue}{\textbf{23.35}}&6.81&\textcolor{Blue}{\textbf{24.11}}&\textcolor{Blue}{\textbf{28.67}} \\ 
\textbf{Ours}* & \textcolor{Red}{\textbf{4.46}}&\textcolor{Red}{\textbf{25.24}}&\textcolor{Blue}{\textbf{5.82}}&\textcolor{Red}{\textbf{33.66}}&\textcolor{Red}{\textbf{0.63}}&\textcolor{Red}{\textbf{10.12}}&\textcolor{Red}{\textbf{28.87}}\\ 
     \bottomrule
     \bottomrule
\end{tabular}
\begin{tablenotes}
\footnotesize
\item[*] with blending operation
\end{tablenotes}
}
\end{threeparttable}
}
}
\label{tab:editbench}
\vspace{-0.6cm}
\end{table}

The performance on the EditBench is roughly consistent with the overall performance on \Benchmark, which similarly shows \OurMethod's superior performance. 
This indicates that our method exhibits strong performance across a range of inpainting tasks with various mask types, including random masks, inside-inpainting masks, and outside-inpainting masks.

\vspace{-0.4cm}

\subsection{Qualitative Comparison}

\vspace{-0.15cm}

The qualitative comparison with previous image inpainting methods is shown in Fig.~\ref{fig:teaser}. 
We provide results on artificial images and natural images across various inpainting tasks, including random mask inpainting, segmentation mask inside-inpainting, and segmentation mask outside-inpainting. 
\OurMethod~consistently show exceptional results in the coherent of generated region and unmasked region, considering content (\uppercase\expandafter{\romannumeral1}, \uppercase\expandafter{\romannumeral2} right, \uppercase\expandafter{\romannumeral3} right, \uppercase\expandafter{\romannumeral4}), color (\uppercase\expandafter{\romannumeral2} left), and text (\uppercase\expandafter{\romannumeral3} left).
Interestingly, Fig.~\ref{fig:teaser} \uppercase\expandafter{\romannumeral3} left requires the model to generate a cat and a goldfish.
All previous methods fail to recognize that a goldfish is already present in the masked image, resulting in the generation of an additional fish within the masked region.
\OurMethod~successfully realized the awareness of background information due to the design of the dual-branch decoupling.

\begin{figure}[htbp]
\vspace{-0.6cm}
    \centering
    \includegraphics[width=0.99\linewidth]{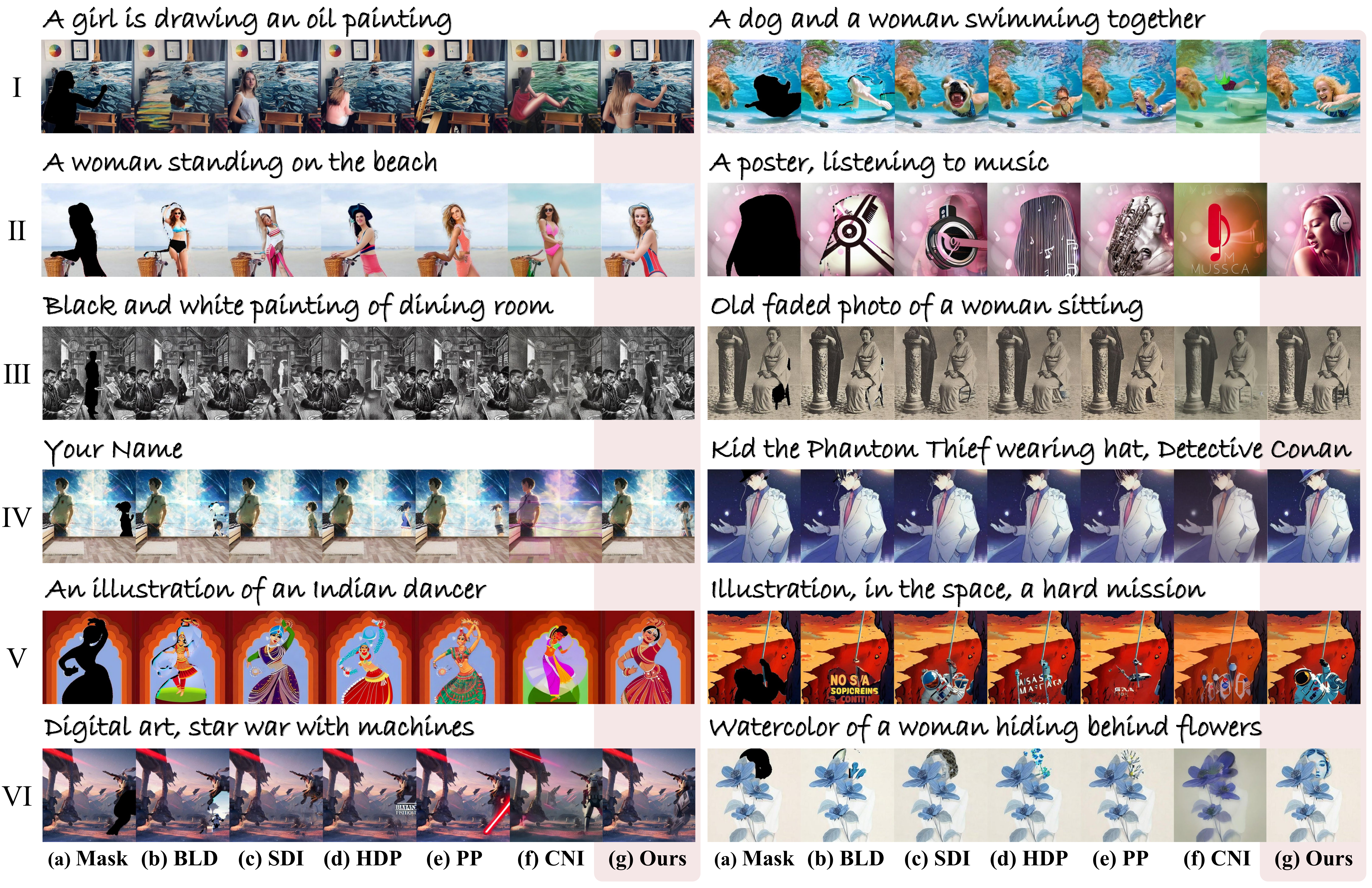}
\vspace{-0.2cm}
    \caption{\textbf{Comparison of previous inpainting methods and \OurMethod~on various image domain.} A detailed explanation of compared methods is in Fig.~\ref{fig:teaser}.}
    \label{fig:experiment_compare}
\vspace{-0.7cm}
\end{figure}

The untouched pre-trained diffusion branch also provides the advantage of better coverage across different data domains, such as painting and anime. As shown in Fig.~\ref{fig:experiment_compare}, \OurMethod~demonstrates superior performance across various image categories, including natural image (\uppercase\expandafter{\romannumeral1}, \uppercase\expandafter{\romannumeral2}), pencil painting (\uppercase\expandafter{\romannumeral3}), anime (\uppercase\expandafter{\romannumeral4}), illustration (\uppercase\expandafter{\romannumeral5}), digital art (\uppercase\expandafter{\romannumeral6} left), and watercolor (\uppercase\expandafter{\romannumeral6} right). Due to the page limit, more qualitative comparison results are in supplementary files.

\vspace{-0.4cm}

\subsection{Flexible Control Ability}
\label{sec:flexible_control_ability}

\begin{figure}[htbp]
    \centering
    \includegraphics[width=1.\linewidth]{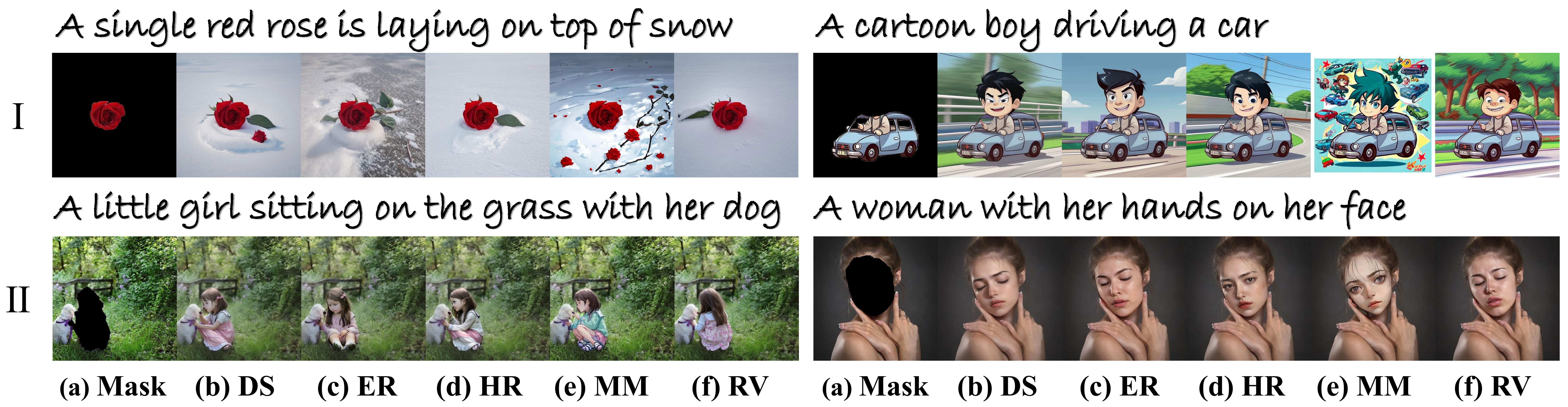}
    \caption{\textbf{Integrating \OurMethod~to community fine-tuned diffusion models.} We use five popular community diffusion models fine-tuned from stable diffusion v1.5: DreamShaper (DS)~\cite{dreamshaper}, epiCRealism (ER)~\cite{epiCRealism}, Henmix\_Real (HR)~\cite{HenmixReal}, MeinaMix (MM)~\cite{MeinaMix}, and Realistic Vision (RV)~\cite{RealisticVision}. MM is specifically designed for anime images.}
    \label{fig:base_model}
\end{figure}

\begin{figure}[htbp]
    \centering
    \includegraphics[width=1.\linewidth]{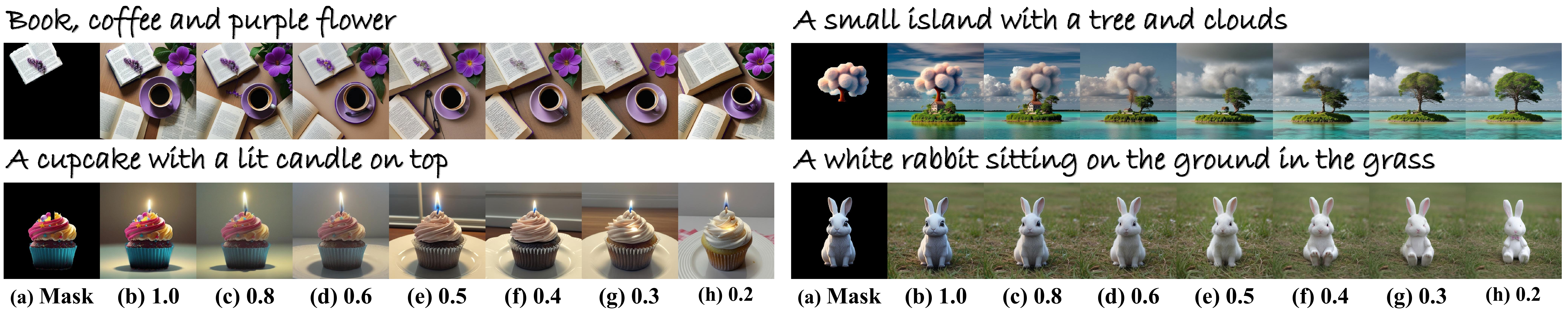}
    \caption{\textbf{Flexible control scale of \OurMethod.} (a) shows the given masked image, (b)-(h) show adding \OurMethod~with control scale $w$ from $1.0$ to $0.2$. Results show a gradually diminishing controllable ability from precise to rough control.}
    \label{fig:scale}
\vspace{-0.4cm}
\end{figure}

Fig.~\ref{fig:base_model} and Fig.~\ref{fig:scale} illustrate the flexible control provided by \OurMethod~in two aspects: base diffusion model selection and controlling scale. In Fig.~\ref{fig:base_model}, we showcase the ability to combine \OurMethod~with different diffusion models fine-tuned by the community. This allows users to select a specific model that best suits their inpainting requirements, enabling users to achieve the desired inpainting results based on their specific needs. Fig.~\ref{fig:scale} demonstrates the adjustment of the control scale of \OurMethod. This control scale parameter allows users to effectively control the extent of unmasked region protection during the inpainting process. By manipulating the scale parameter, users can achieve fine-grained control over the inpainting process, enabling precise and customizable inpainting.

\vspace{-0.4cm}

\subsection{Ablation Study}

\begin{table}[htbp]
\vspace{-0.8cm}
\centering
\scriptsize
\caption{\textbf{Ablation on dual-branch design.} Stable Diffusion Inpainting (SDI) use single-branch design, where the entire UNet is fine-tuned. We conducted an ablation analysis by training a dual-branch model with two variations: one with the base UNet fine-tuned, and another with the base UNet forzened. Results demonstrate the superior performance achieved by adopting the dual-branch design. \textcolor{Red}{\textbf{Red}} is the best result.}
\vspace{-0.15cm}
\scalebox{1.0}{
\setlength{\tabcolsep}{0.9mm}{
\begin{threeparttable}
{
\begin{tabular}{c|ccc|ccc|c}
\toprule
\toprule
\multicolumn{1}{c|}{\textbf{Metrics}}  & \multicolumn{3}{c|}{\textbf{Image Quality}} & \multicolumn{3}{c|}{\textbf{Masked Region Preservation}} & \textbf{Text Align} \\
\midrule
\textbf{Model} & \textbf{IR}$_{^{\times 10}}$$\uparrow$   & \textbf{HPS}$_{^{\times 10^2}}$$\uparrow$  & \textbf{AS}$\uparrow$   & \textbf{PSNR}$\uparrow$     & \textbf{LPIPS}$_{^{\times 10^2}}$$\downarrow$  & \textbf{MSE}$_{^{\times 10^3}}$$\downarrow$      & \textbf{CLIP Sim}$\uparrow$     \\ \midrule
SDI & 11.00 & 27.53 & 6.53 & 19.78 & 16.87 & 31.76 & 26.69 \\ 
w/o fine-tune & 11.59 & 27.71 & 6.59 & 19.86 & 16.09 & 31.68 & 26.91 \\ 
w/ fine-tune & \textcolor{Red}{\textbf{11.63}} & \textcolor{Red}{\textbf{27.73}} & \textcolor{Red}{\textbf{6.60}} & \textcolor{Red}{\textbf{20.13}} & \textcolor{Red}{\textbf{15.84}} & \textcolor{Red}{\textbf{31.57}} & \textcolor{Red}{\textbf{26.93}} \\ 

     \bottomrule
     \bottomrule
\end{tabular}
}
\end{threeparttable}
}
}
\vspace{-0.5cm}
\label{tab:ablation_finetune}
\end{table}

We conducted ablation studies to investigate the impact of different model designs. Tab.~\ref{tab:ablation_finetune} compares the dual-branch and single-branch designs. Tab.~\ref{tab:ablation} shows the ablation study focusing on the additional branch architecture.
The ablation studies are conducted on \Benchmark, averaging the performance of inside-inpainting and outside-inpainting.
The results presented in Tab.~\ref{tab:ablation_finetune} demonstrate that the dual-branch design significantly outperforms the single-branch design. Additionally, fine-tuning the base diffusion model in dual-branch design yields better results compared to freezing it.
However, fine-tuning the base diffusion model may restrict flexibility and control over the model. 
Considering this trade-off between performance and flexibility, we decide to adopt the frozen dual-branch design as our model design. 
Tab.~\ref{tab:ablation} presents the rationale behind the design choices for (1) using a VAE encoder instead of randomly initialized convolution layers to process the masked image, 
(2) incorporating the full UNet feature layer-by-layer into the pre-trained UNet, 
(3) removing text cross-attention in \OurMethod, which avoids masked image features influenced by text.

\begin{table}[htbp]
\vspace{-0.5cm}
\centering
\scriptsize
\caption{\textbf{Ablation on model architecture.} 
We ablate on the following components: the image encoder (Enc), selected from a random initialized convolution (Conv) and a VAE; the inclusion of mask in input (Mask), chosen from adding (w/) and not adding (w/o); the presence of cross-attention layers (Attn), chosen from adding (w/) and not adding (w/o); the type of UNet feature addition (UNet), selected from adding the full UNet feature (full), adding half of the UNet feature (half), and adding the feature like ControlNet (CN); and finally, the blending operation (Blend), chosen from not adding (w/o), direct pasting (paste), and blurred blending (blur). \textcolor{Red}{\textbf{Red}} is the best result.}
\vspace{-0.1cm}
\scalebox{0.85}{
\setlength{\tabcolsep}{0.5mm}{
\begin{threeparttable}
{
\begin{tabular}{ccccc|ccc|ccc|c}
\toprule
\toprule
\multicolumn{5}{c|}{\textbf{Metrics}}  & \multicolumn{3}{c|}{\textbf{Image Quality}} & \multicolumn{3}{c|}{\textbf{Masked Region Preservation}} & \textbf{Text Align} \\
\midrule
\textbf{Enc} & \textbf{Mask} & \textbf{Attn} & \textbf{UNet} & \textbf{Blend} & \textbf{IR}$_{^{\times 10}}$$\uparrow$   & \textbf{HPS}$_{^{\times 10^2}}$$\uparrow$  & \textbf{AS}$\uparrow$   & \textbf{PSNR}$\uparrow$     & \textbf{LPIPS}$_{^{\times 10^2}}$$\downarrow$  & \textbf{MSE}$_{^{\times 10^3}}$$\downarrow$      & \textbf{CLIP Sim}$\uparrow$     \\ \midrule 
\cellcolor{LightRed} Conv & w/ & w/o & full & \cellcolor{LightRed} w/o & 11.05&26.23&6.55&14.89&37.23&64.54&26.76\\
VAE & \cellcolor{LightRed} w/o & w/o & full & \cellcolor{LightRed} w/o& 11.55 & 27.70 & 6.57 & 17.96 & 26.38 & 49.33 & 26.87\\
VAE & w/ & \cellcolor{LightRed} w/ & full & \cellcolor{LightRed} w/o& 11.25 & 27.62 & 6.56 & 18.69 & 19.44 & 34.28 & 26.63\\
\cellcolor{LightRed} Conv & w/ & \cellcolor{LightRed} w/ & \cellcolor{LightRed} CN & \cellcolor{LightRed} w/o&  9.58&26.85&6.47&12.15 & 80.91 & 150.89&26.88\\
VAE & w/ & \cellcolor{LightRed} w/ & \cellcolor{LightRed} CN & \cellcolor{LightRed} w/o& 10.53 & 27.42 & 6.59 & 18.28 & 24.36 & 41.63 & 26.89\\
VAE & w/ & w/o & \cellcolor{LightRed} CN & \cellcolor{LightRed} w/o& 11.42 & 27.69 & 6.58 & 18.49 & 24.09 & 36.33 & 26.86 \\
VAE & w/ & w/o & \cellcolor{LightRed} half & \cellcolor{LightRed} w/o& 11.47 & 27.70 & 6.57 & 19.01 & 23.77 & 33.57 & 26.87 \\
VAE & w/ & w/o & full & \cellcolor{LightRed} w/o & 11.59 & 27.71 & \textcolor{Red}{\textbf{6.59}} & 19.86 & 16.09 & 31.68 & \textcolor{Red}{\textbf{26.91}}\\
VAE & w/ & w/o & full & \cellcolor{LightRed} 
 paste &11.72&27.93&6.58&-&-&-&26.80\\
\midrule 
VAE & w/ & w/o & full & blur & \textcolor{Red}{\textbf{11.76}}&\textcolor{Red}{\textbf{27.94}}&6.58&\textcolor{Red}{\textbf{29.88}}&\textcolor{Red}{\textbf{1.53}}&\textcolor{Red}{\textbf{11.65}}&26.81 \\ 

     \bottomrule
     \bottomrule
\end{tabular}
}
\end{threeparttable}
}
}
\label{tab:ablation}
\vspace{-0.5cm}
\end{table}

\vspace{-0.45cm}
\section{Discussion}
\label{sec:conclusion}

\vspace{-0.25cm}

\paragraph{\textbf{Conclusion.}} This paper proposes a plug-and-play image inpainting method \OurMethod~with a pixel-level masked image feature insertion architectural design. Quantitative and qualitative results on our proposed benchmark, \Benchmark, and EditBench show the superior performance of \OurMethod~considering image generation quality, masked region preservation, and image-text alignment. 

\vspace{-0.2cm}

\paragraph{\textbf{Limitations and Future Work.}} However, \OurMethod~still has some limitations: (1) The quality and content generated by our model are heavily dependent on the chosen base model. As shown in Figure~\ref{fig:base_model}, the results of Model MeinaMix~\cite{MeinaMix} exhibit incoherence because the given image is a natural image while the generation model primarily focuses on anime. (2) Even with \OurMethod, we still observe poor generation results in cases where the given mask has an unusually shaped or irregular form, or when the given text does not align well with the masked image. In our future work, we will continue to address these challenges and further improve upon the identified problems.

\vspace{-0.2cm}

\paragraph{\textbf{Negative Social Impact.}}
Image inpainting models present exciting opportunities for content creation, but they also carry potential risks to individuals and society. Their reliance on internet-collected training data can amplify social biases, and there is a specific risk of generating persuasive misinformation by manipulating human images with offensive elements. To address these concerns, it is crucial to emphasize responsible use and establish ethical guidelines when utilizing these models. This is also a key focus for our future model releases.

\bibliographystyle{splncs04}
\bibliography{egbib}
\end{document}